\documentclass[draft]{ecai} 

\usepackage{latexsym}
\usepackage{amssymb}
\usepackage{amsmath}
\usepackage{amsthm}
\usepackage{algorithm}
\usepackage{algorithmicx}
\usepackage{algpseudocode}
\usepackage{booktabs}
\usepackage{enumitem}
\usepackage[final]{graphicx}
\usepackage{color}
\usepackage[caption=false]{subfig}
\usepackage{tabularx}
\usepackage{multirow}
\usepackage{tcolorbox}

\usepackage[final]{listings}
\usepackage{amsfonts}
\usepackage{xcolor}
\usepackage{siunitx}
\lstdefinelanguage{Prolog}{
    morekeywords={:-, ?, is, not, fail, true, listing, and, if},
    sensitive=true,
    morecomment=[l]{\%},
    morestring=[b]",
}

\definecolor{darkblue}{RGB}{0,0,139}

\newcommand{\gpt}{GPT-4o}
\newcommand{\claude}{Claude 3.5 Sonnet}

\begin{document}

\begin{frontmatter}


\title{Generative Agents for Multi-Agent Autoformalization of Interaction Scenarios}

\author[A]{\fnms{Agnieszka}~\snm{Mensfelt}\orcid{0000-0002-2385-2017}}
\author[A]{\fnms{Kostas}~\snm{Stathis}\orcid{0000-0002-9946-4037}}
\author[A]{\fnms{Vince}~\snm{Trencsenyi}\orcid{0009-0009-4560-7571}} 

\address[A]{Royal Holloway, University of London, Egham, Surrey, UK}


\begin{abstract}
Multi-agent simulations are versatile tools for exploring interactions among natural and artificial agents, but their development typically demands domain expertise and manual effort. This work introduces the Generative Agents for Multi-Agent Autoformalization (GAMA) framework, which automates the formalization of interaction scenarios in simulations using agents augmented with large language models (LLMs). To demonstrate the application of GAMA, we use natural language descriptions of game-theoretic scenarios representing social interactions, and we autoformalize them into executable logic programs defining game rules, with syntactic correctness enforced through a solver-based validation. To ensure runtime validity, an iterative, tournament-based procedure tests the generated rules and strategies, followed by exact semantic validation when ground truth outcomes are available. In experiments with 110 natural language descriptions across five $2\times2$ simultaneous-move games, GAMA achieves 100\% syntactic and 76.5\% semantic correctness with \claude{}, and 99.82\% syntactic and 77\% semantic correctness with \gpt{}. The framework also shows high semantic accuracy in autoformalizing agents' strategies.

\end{abstract}

\end{frontmatter}


\section{Introduction}

Multi-agent simulations (MAS) provide insights into diverse phenomena, ranging from individual human interactions and organizational dynamics to biological processes, robotics, and interactions among artificial agents. However, modelling real-world scenarios and implementing corresponding simulations typically require substantial human expertise and effort. Recent advances in large language models (LLMs) and autoformalization techniques~\cite{wu2022autoformalization} offer a promising new approach to streamlining these tasks. In this work, we demonstrate this approach using interaction scenarios modelled as $2\times2$ simultaneous-move games.

LLMs have been increasingly explored as agents in interaction scenarios and game-theoretic contexts, though often under a weak notion of agency~\cite{akata2023playing,fan2023can,guo2023gpt,lore2023strategic}. Furthermore, their inherent limitations—such as hallucinations~\cite{achiam2023gpt} and logical or arithmetic errors~\cite{imani2023mathprompter}—undermine their reliability as rational decision-makers in simulations. Addressing this challenge, autoformalization shifts the role of LLMs from decision-making to format translation, converting natural language interaction descriptions into formal representations. This leverages one of LLMs' core strengths, with autoformalization showing considerable promise in formalizing mathematics~\cite{wu2022autoformalization,he2023solving,jiang2022draft} and logic~\cite{yang2023coupling,pan2023logic,cosler2023nl2spec,chen2023nl2tl,feng2023language}.

In multi-agent systems, such formalized interaction descriptions can aid and, often act, as decision-making modules, reducing reliance on the LLM’s reasoning correctness during runtime. Moreover, this formal representation can be iteratively refined using automatic reflection and refinement mechanisms~\cite{madaan2023selfrefine,renze2024self,shinn2023reflexion,gou2024critic}, enabling continuous improvement via feedback loops. This shift from natural language reasoning to formal, verifiable representations establishes a more robust foundation for automating LLM-driven agent design.

Our prior work~\cite{mensfelt24a}, introduced an autoformalization module capable of converting interaction descriptions into syntactically validated code. The Generative Agents for Multi-Agent Autoformalization (GAMA) incorporates the previous work as a component of the simulation framework, while additionally providing runtime and semantic validation and enabling the generation of formal representations of both strategic interactions and game strategies. These formalizations act as reasoning modules for agents deployed in simulations, enabling comparison of different strategies for given interactions.

The main contributions of this work are as follows:
\begin{enumerate}[label=\Roman*.]
\item \textbf{Agent model with formal game and strategy representations}: We propose an agent architecture that integrates formalized representations of games and strategies alongside the autoformalization module. Game rules and strategies can be autoformalized or supplied as predefined specifications.
    
\item \textbf{Three-level agent validation}: We introduce a multi-level validation pipeline for agents, including (1) syntactic validation using formal solvers, (2) runtime validation through simulation-based testing, and (3) semantic validation based on ground truth outcomes (when available) or compliance with game-specific constraints.

\item \textbf{Simulation framework for autoformalized scenarios}: We present a simulation environment that supports automated validation and facilitates the exploration and comparison of strategies formalized from natural language descriptions.

\item \textbf{Experimental evaluation}: We empirically evaluate GAMA, demonstrating high syntactic and semantic correctness across 110 scenarios modelled as $2\times2$ simultaneous-move games. Furthermore, we assess the agents' ability to autoformalize and execute five distinct gameplay strategies.

\end{enumerate}

The remainder of this paper is organized as follows. Section~\ref{sec:preliminaries} introduces the background and preliminaries. Section~\ref{sec:framework} presents the GAMA framework. Section~\ref{sec:methods} details the experimental setup and evaluation methodology. Section~\ref{sec:results} reports the experimental results. Finally, Section~\ref{sec:conclusions} concludes the paper and outlines directions for future work.

\section{Preliminaries}
\label{sec:preliminaries}

\subsection{Game Theory}
\label{sec:game-theory}

Game theory is an analytical framework for strategic interactions, where games are composed of four main components~\cite{Rasmusen2004introtogametheory}:
\begin{description}
    \item[\textbf{Players}] players consistently and willingly act towards maximising their utility, in full awareness of the situation they are involved in.
    
    \item[\textbf{Actions}] mappings between states and outcomes, carried out by players \cite{OsborneRubinstein1994}.
    
    \item[\textbf{Payoffs}] payoffs are quantitative measures over the consequent outcomes of action tuples chosen by players \cite{Gibbons1992}.
    
    \item[\textbf{Information}] game information describes a player's knowledge about the available actions, historical moves, and the player's image of its opponents \cite{OsborneRubinstein1994}.
\end{description}

We formally define games in this context as a tuple $(N, A, \Pi)$ consisting of a set of $n$ players $N$, a non-empty set of actions $A_i,\forall i \in N$, and for each player $i$ a payoff function $\pi_i: \times_{j \in N} A_j \rightarrow \mathbb{R}$ mapping action profiles to player payoffs.

\begin{table}[h]
    \caption{A general normal form game describing the four possible outcomes $W,X,Y,Z$, where the tuples denote the row player's and column player's corresponding payoffs, in general: $(\pi_{row}, \pi_{col})$.}
    \centering
    \begin{tabular}{rcc}
        \toprule
        \textbf{row/col} & \textbf{L} & \textbf{R} \\
        \midrule
        \textbf{U} & ($W_{row}$, $W_{col}$) & ($X_{row}$, $X_{col}$) \\
        \textbf{D} & ($Y_{row}$, $Y_{col}$) & ($Z_{row}$, $Z_{col}$) \\
        \bottomrule
    \end{tabular}
    \label{tab:pd}
\end{table}

In the case of symmetric games, the payoffs solely depend on the chosen strategies, while with asymmetric games, the outcomes yield different payoffs based on the player's identity. In our experiments, the Prisoner's Dilemma, Hawk-Dove, and Stag-Hunt are symmetric, and we classify actions U and L as ``Cooperate'' denoted by $C$ and actions D and R as ``Defect'' denoted by $D$. This allows us to use the terminology from Axelrod's tournaments~\cite{axelrod1981evolution} to define these games in terms of the four outcomes: $\mathbf{T}:(D,C),\mathbf{R}:(C,C),\mathbf{P}:(D,D),\mathbf{S}:(C,D)$.

\begin{description}
    \item[\textbf{Prisoner's Dilemma}] is characterised by a dominant temptation to defect, represented by the payoff relation $T > R > P > S$.
    
    \item[\textbf{Hawk-Dove}] entails switched relative payoffs between the punishment for mutual defection and the sucker's payoff: $T > R > S > P$.
    
    \item[\textbf{Stag Hunt}] involves a relatively high reward for mutual cooperation over the temptation to defect: $R > T > P > S$.
\end{description}

The Battle of the Sexes and Matching Pennies are asymmetric games. As the outcomes mean different payoffs for \textit{row} and \textit{column}, we cannot use the generalization from above and have to define the games from both players' perspectives:

\begin{description}
    \item[\textbf{Battle of the Sexes}] is a coordination game, where both players prefer an agreement but have different preferences over the coordinated outcomes:
    \begin{itemize}
        \item \textit{Row}: $W > Z > \{X, Y\}$.
        \item \textit{Column}: $Z > W > \{X, Y\}$.
    \end{itemize}

    \item[\textbf{Matching Pennies}] coordinated outcomes reward the row player, while mismatched actions benefit the column player:
    \begin{itemize}
        \item \textit{Row}: $\{W, Z\} > \{X, Y\}$.
        \item \textit{Column}: $\{X, Y\} > \{W, Z\}$.
    \end{itemize}
\end{description}

\subsection{Large Language Models}

The rapid advancement of natural language processing (NLP), driven by transformer architectures~\cite{gillioz2020transformers} and pre-trained models~\cite{Qiu2020}, has led to the emergence of Large Language Models. State-of-the-art (SOTA) LLMs continue to evolve rapidly, primarily due to the increasing scale of model parameters and training datasets~\cite{zhao2023survey}. 

In this work, we employ OpenAI’s GPT-4 Omni~\cite{gpt4o2024}, a multimodal SOTA model capable of processing text, images, and audio. It has demonstrated strong performance on reasoning benchmarks~\cite{wang2024rupbenchbenchmarkingreasoningperturbations}. Likewise, we utilize Anthropic’s Claude 3.5 Sonnet, another high-performing SOTA model, which excels across multiple benchmarks~\cite{claude35}.

\subsection{General Game Playing}
General game playing~\cite{gdl} focuses on developing intelligent systems that explicitly represent the rules of arbitrary new games and learn to play them autonomously, without human intervention. The Game Description Language (GDL) has been proposed as a formal, machine-processable language for describing the rules of arbitrary games~\cite{gdl-orig}. GDL focused on information games only, so it was extended in GDL-II~\cite{gdl-ii} to cover n-player games with incomplete information and games in extensive normal form~\cite{gdl-univ}.

A challenge with GDL systems is that learning without human guidance demands complex reasoning. Players must infer the possible actions of others, effectively analysing hypothetical game scenarios before making decisions. Action formalisms like the classical Situation Calculus~\cite{sc} have been developed for precisely this purpose. Formal schemes and inference methods are readily available for Situation Calculus~\cite{sc-games1,sc-games2}, while their deployment in general game playing presupposes a  translation from GDL into existing, suitably expressive action languages. One such scheme~\cite{Schiffel_Thielscher_2011} shows how to fully embed GDL-II into a version of the Situation Calculus based on knowledge fluents~\cite{sc-kbf}. Our game solver is inspired by these previous works, to support light GDL specifications that act as target representation for the autoformalization.

\section{Generative Agents for Multi-Agent Autoformalization}
\label{sec:framework}

\subsection{Overview}

\begin{figure}[htb]
    \centering
    \includegraphics[width=\linewidth]{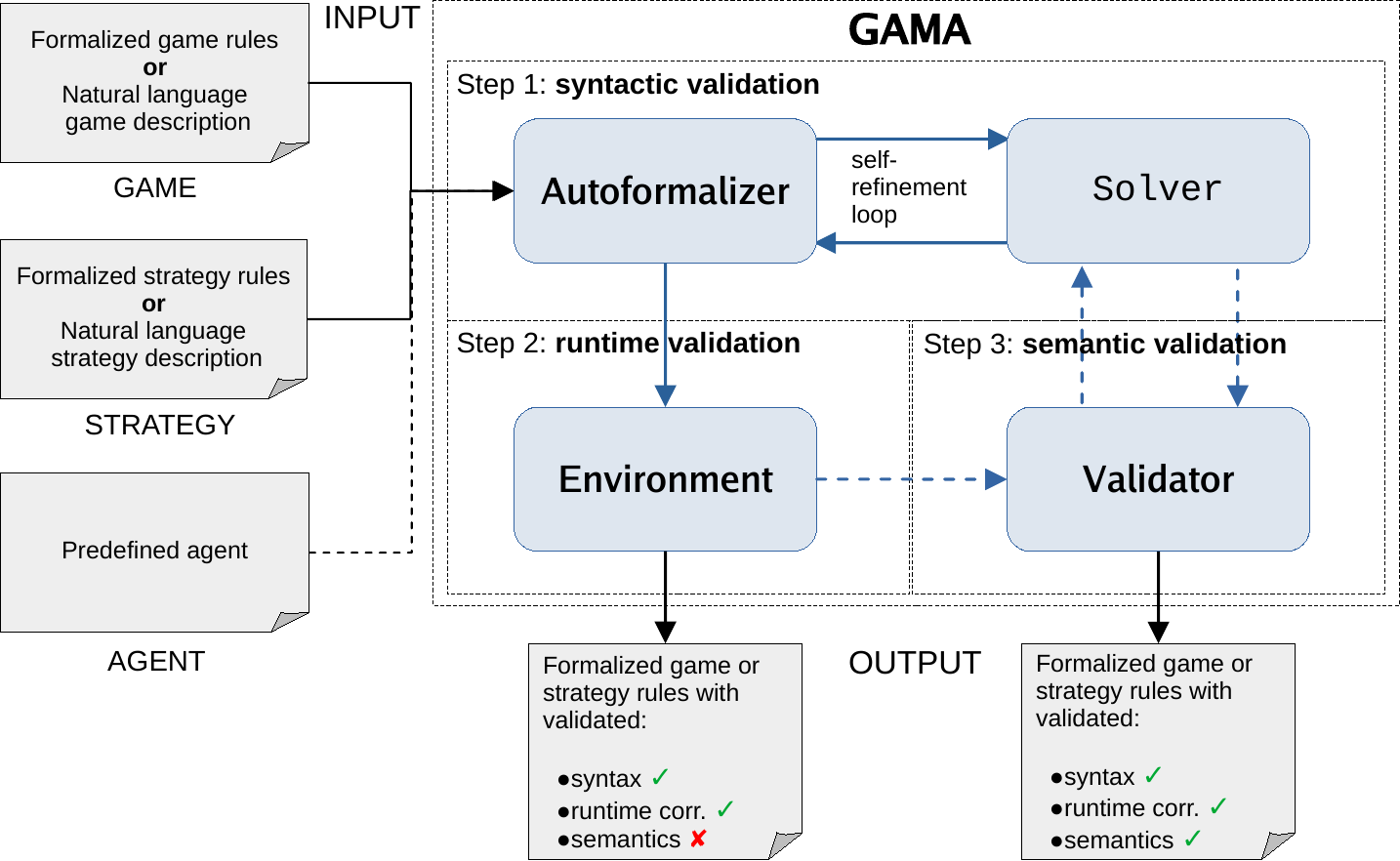}

\caption{Overview of the GAMA framework. Dashed lines indicate optional control flows. Games and strategies can be specified either through predefined rules or natural language descriptions, or optionally by loading a predefined agent. Semantic validation is performed when ground truth outcomes are available. The Validator may employ a formal solver.}

    \label{fig:framework-schema}
\end{figure} 

Fig.\ref{fig:framework-schema} presents an overview of the GAMA framework. The framework is implemented in Python and Prolog~\cite{wielemaker2012swi}, though its modular architecture supports substituting alternative solvers. The source code and log files generated during the evaluation are available at\footnote{\textcolor{blue}{\url{https://github.com/dicelab-rhul/GAMA}}}.

Each agent comprises an autoformalization module and a logic programming component, which is further divided into two modules: a \textit{Game} module, containing both game-independent and game-specific rules, and a \textit{Strategy} module, which specifies the agent's action selection logic against an opponent. Together, these modules form a solver that encapsulates the game logic and determines the agent's next move. A Python wrapper enables interaction with the external environment, manages game history (including each agent's moves, opponent actions, and payoffs), and updates the solver’s internal state accordingly.

\subsection{Agents' Initialization and Syntactic Validation}

\begin{algorithm}
	\small
	\caption{Generating game rules predicates from natural language descriptions. PD stands for Prisoner's Dilemma.}
	\begin{algorithmic}[1]
		\State \textbf{Input:} \newline \texttt{$\Gamma$}: game-independent predicates, \newline \texttt{$NL_{PD}$}: natural language description of PD, \newline \texttt{$\xi_{PD}$}: game-specific predicates of PD, \newline \texttt{$NL_{NG}$}: natural language description of a new game. 
		\State \textbf{Output:} \newline \texttt{$\xi_{NG}$}: game-specific predicates for the new game. 
		\State \textbf{Parameter:} \newline \texttt{$\text{max\_attempts}$}: maximum correction attempts.
		
		\State \texttt{$\text{attempts} \gets 0$}
		\State \texttt{$\text{trace} \gets \emptyset$}
		
		\While{\texttt{$\text{attempts} < \text{max\_attempts}$}}
		\State \texttt{$\xi_{NG} \gets \text{LLM.translate}(\Gamma, NL_{PD}, \xi_{PD}, NL_{NG})$}
		\State \texttt{$\text{is\_valid} \gets \text{solver.check\_predicates}(\xi_{NG})$}
		
		\If{\texttt{$\text{is\_valid}$}}
		\State \Return \texttt{$\xi_{NG}$}
		\Else
		\State \texttt{$\text{trace} \gets \text{solver.get\_trace}()$}
		\State \texttt{$\xi_{NG} \gets \text{LLM.self\_correct}(\xi_{NG}, \text{trace})$}
		\EndIf
		\State \texttt{$\text{attempts} \gets \text{attempts} + 1$}
		\EndWhile
		
		\State \Return \texttt{Unable to generate valid predicates within maximum attempts.}	
	\end{algorithmic}
	\label{alg:algorithm}
\end{algorithm}

To perform autoformalization, an agent employs a large language model (LLM) through a unified abstract interface, allowing compatibility with any LLM. Natural language descriptions of games or strategies are formalized using one-shot prompting, with examples of game-dependent predicates (e.g., for the Prisoner's Dilemma or tit-for-tat strategy) provided in the prompt. Once the code is generated, its syntactic correctness is validated. If errors are detected, the LLM receives feedback containing the erroneous lines, error messages, and instructions for correcting the predicates. The process is repeated up to a limit defined by the \texttt{max\_attempts} parameter. An overview of the autoformalization module~\cite{mensfelt24a} is shown in Listing~\ref{alg:algorithm}.

\subsection{Simulation and Runtime Validation}

After generating a pool of syntactically correct agents, the framework operates in two primary modes: validation and strategy evaluation. These modes enable assessment of both the runtime correctness of generated programs and the effectiveness of strategies across various game scenarios.

The key parameters for tournament-based simulation are as follows:
\begin{description}
\item[\textbf{Agent Pool}] All participating agents, initialized with game rules and strategies obtained either through autoformalization or predefined configurations.

\item[\textbf{Number of Rounds}] The number of rounds each agent pair will play.

\item[\textbf{Match Maker}] A customizable function that defines how agents are paired for matches.

\item[\textbf{Target Payoffs}] (Optional) Used for semantic validation when ground truth payoffs are available.
\end{description}

Tournament outcomes can be determined in two ways:
\begin{enumerate}
\item Selecting agents that achieve the highest total payoff, reflecting strategy effectiveness.
\item Selecting agents whose total payoff matches the target payoff (if available), used primarily for validation.
\end{enumerate}

In \textit{strategy evaluation mode}, the framework compares strategies based on their performance within game scenarios. Agents equipped with different strategies typically compete in a round-robin tournament, where each agent plays against every other agent for a specified number of rounds. However, alternative competition formats can be defined using a custom \textbf{Match Maker} function. By simulating multiple matches, the framework identifies which strategies perform best on average, providing insights into their relative strengths and weaknesses.

\subsection{Semantic Validation}

Semantic validation is performed when ground truth outcomes are available. Here, “semantic correctness” refers to the correspondence between the generated code and the input natural language description. For example, formalized code for the Prisoner's Dilemma that is syntactically valid and executes correctly in simulation, but encodes an incorrect payoff matrix, would be considered semantically incorrect.

In \textit{semantic validation mode}, when specific target outcomes are provided, the framework performs exact validation. Once a scenario is autoformalized and an agent is instantiated with the given strategy, the agent plays against a clone of itself for a fixed number of rounds. The resulting outcomes for all action tuples are then compared directly to the target outcomes to assess semantic correctness.

If specific target outcomes are not available but the game type is known (e.g., it is known that the scenario represents a Prisoner's Dilemma, but the exact payoff matrix is unknown), the framework performs constraint-based validation. In this case, it verifies that the generated payoff matrix satisfies all necessary properties (such as symmetry or payoff dominance relations) defined for that game type (Section~\ref{sec:val-constraints}). This ensures that the generated logic faithfully captures the intended structure of the game.

\subsection{Solver}
\label{sec:solver}

\label{subsec:solver}
Our solver is based on logic programming, as described in~\cite{mensfelt24a}, but repeated here for completeness. It comprises of a game-independent part representing the rules for any normal-form game, a game-dependent part defining the rules of a particular game using the predicates from the game-independent part, and a set of auxiliary predicates completing the definition of that particular game. We follow standard Prolog conventions to interpret a game: variables are indicated by uppercase letters, while predicates and function symbols by lowercase. 
An underscore \texttt{'\_'} represents a variable whose value is ignored within a definition. 
The game state is then a situation, initially denoted by a constant (e.g. {\tt s0}). The binary function {\tt do(M, S)} denotes the situation resulting from the executing move {\tt M} in situation {\tt S}, consistent with the Situation Calculus.
\subsubsection{Game-independent part}
We define all legal transitions of an extended-form game from an initial situation {\tt S} to a final situation {\tt F} as follows:
\begin{lstlisting}
game(F, F) if 
    final(F).

game(S, F) if 
    not final(S) and 
    legal(M, S) and 
    game(do(M, S), F).
\end{lstlisting}
The game terminates when the final game situation {\tt F} is reached. Otherwise, in a non-final situation {\tt S}, the game accepts a legal move {\tt M}, and the game continues in the next {\tt do(M,S)} situation, until the final situation {\tt F} is reached. To determine what holds in each legal situation, we use Situation Calculus, represented as:

\begin{lstlisting}
holds(F, S) if 
    initially(F, S).

holds(F, do(M, S)) if 
    effect(F, M, S).

holds(F, do(M, S)) if 
    holds(F, S) and 
    not abnormal(F, M, S).
\end{lstlisting}
{\tt holds/2} here is similar to {\tt true/1} in GDL, but with an additional parameter of the situation in which the fluent is true. It states that a fluent {\tt F} holds in the initial situation ({\tt init/1} in GDL), a new fluent {\tt F} is initiated by the effects of a move {\tt M} executed in a situation {\tt S} ({\tt next/1} in GDL), and a fluent {\tt F} persists after a move is made, provided it is not abnormal; abnormal fluents do not persist (implicit in GDL). We also use rules of the form:
\begin{lstlisting}
finally(F, S) if Conditions.
\end{lstlisting}
to return derived fluents {\tt F} describing the result of the game, when the {\tt Conditions} hold  in the final situation {\tt S}. 

\subsubsection{Game-dependent part} To represent a specific game we need to define game-dependent predicates for the initial state {\tt initial/1}, the legal moves {\tt legal/2}, what holds in the initial game situation via {\tt initially/2}, the effects of a move on a situation via {\tt effect/3}, what stops persisting in a situation after the execution of a move via {\tt abnormal/3}, the final situation {\tt final/1}, and the result of the game via {\tt finally/2} definitions. To exemplify these definitions, we show how to describe a PD game to our solver.
The initial situation {\tt s0} is defined as:
\begin{lstlisting}
initial(s0).
\end{lstlisting}
What holds in this initial situation we specify it as:
\begin{lstlisting}
initially(player(p1), s0).
initially(player(p2), s0).
initially(role(p1,row), s0).
initially(role(p2,col), s0).
initially(control(p1), s0).
initially(control(p2), s0).
\end{lstlisting}
These assertions define first the player names represented by unique identifiers ({\tt p1} and {\tt p2}), their roles ({\tt p1} is the {\tt row} player, while {\tt p2} is the {\tt col}umn player), and then the fact that initially either of them can the game next (via the {\tt control/1} fluent,as in GDL). What holds in the initial situation changes as a result of move being made in the game. We represent moves as terms of the form {\tt move(P, M)}, where {\tt P} is a player, and {\tt M} is a move. As in PD it is possible for any player to defect ({\tt 'D'}) or cooperate ({\tt 'C'}), we express this as:
\begin{lstlisting}
possible(move(P,'D'),S) if holds(player(P),S).
possible(move(P,'C'),S) if holds(player(P),S).
\end{lstlisting}
It is then legal for a player to make a possible move if they have the control to execute it in the current situation:
\begin{lstlisting}
legal(move(P, M), S) if  
    possible(move(P, M), S) and 
    holds(control(P), S).
\end{lstlisting}
As an effect of a legal move {\tt M} being made by a player {\tt P}, we record in the next situation that the player did that move ({\tt effect/3} is like {\tt next/1} in GDL):
\begin{lstlisting}
effect(did(P, M), move(P, M), S).
\end{lstlisting}
Once a legal move is executed, the player loses control, which we specify in our framework as:
\begin{lstlisting}
abnormal(control(P), move(P, M), S).
\end{lstlisting}
When a player loses control cannot play a move until control is passed back. When each player makes a move from the initial situation, the resulting final situation can be determined as:
\begin{lstlisting}
final(do(M2, do(M1, S))) if
    ground(M2) and 
    ground(M1) and 
    initial(S).
\end{lstlisting}
Assuming the payoff matrix following the structure presented in Table~\ref{tab:pd} defined as:
\begin{lstlisting}
payoff('D', 'D', 1, 1).
payoff('C', 'D', 0, 5).
payoff('D', 'C', 5, 0).
payoff('C', 'C', 3, 3).
\end{lstlisting}
the outcome of the game holds information about the actual moves made by the players, and their payoffs.
\begin{lstlisting}
finally(outcome(P1, M1, U1, P2, M2, U2), S) if
    holds(role(P1, row), S) and 
    holds(did(P1, M1), S) and 
    holds(role(P2, col), S) and 
    holds(did(P2, M2), S) and 
    payoff(M1, M2, U1, U2).
\end{lstlisting}
We can then extract specific outcome information, e.g. the player's utility, through a {\tt goal/2} fluent (as in GDL) e.g.:
\begin{lstlisting}
finally(goal(P1, U1), S) if
    finally(outcome(P1, _, U1, _, _, _), S).

finally(goal(P2, U2), S) if
    finally(outcome(_, _, _, P2, _, U2), S).
\end{lstlisting}
This finalises the definition of PD, enabling us to use the game description above to reason about the game's dynamics. For instance, if player {\tt p1} wanted to analyse the game and identify the actions required to achieve a utility of 5, then it asks the query:
\begin{lstlisting}[xleftmargin=20pt]
?- game(s0,F), finally(goal(p1,5),F).
\end{lstlisting}
Applying the PD payoff matrix, our solver produces two solutions for \texttt{F}, separated by semicolons (;). The result {\tt false} indicates "no further answers":
\begin{lstlisting}[xleftmargin=20pt]
F = do(move(p2,'C'),do(move(p1,'D'),s0));
F = do(move(p1,'D'),do(move(p2,'C'),s0));
false.
\end{lstlisting}
In the first solution, {\tt p1} acts first and {\tt p2} second, while in the second solution, the order is reversed. This outcome is expected, as both players have initial control in our game definition, resulting in both action order combinations being shown.

\subsubsection{Strategy}

Strategies are submitted separately from the game-dependent component.  The \(\texttt{select/4}\) predicate is used to determine the next move {\tt M} of player {\tt P} against opponent {\tt O} in game situation {\tt S}, as illustrated in the \textit{tit-for-tat} strategy example below:

{
\begin{lstlisting}
select(P, O, S, M) if
    not holds(last_move(O, _LMo), S) and
    holds(default_move(P, M), S).
select(P, O, S, Mo) if
    holds(last_move(O, Mo), S).
\end{lstlisting}
}
Here, a player {\tt P} selects the default move first (typically to cooperate), otherwise it mirrors the opponent's move in the last round. {\tt holds/2} is based on the implementation of the Situation Calculus~\cite{sc}. The strategy relies on specific predicates---\(\texttt{default\_move/2}\) and {\tt last\_move/2}---which are required from the game-dependent part of the solver. More sophisticated strategies can be expressed.

\subsubsection{Constraints validation}
\label{sec:val-constraints}

We check that the autoformalized payoff matrix satisfies the structural and content constraints of the specified game (as outlined in Sect.~\ref{sec:game-theory}), by defining validity constraints. The listing below specifies the payoff matrix validity constraints for the Prisoner's Dilemma.
\begin{lstlisting}
    valid_pd_payoffs(T,R,P,S,C,D) if
            payoff(C,C,R,R) and
            payoff(C,D,S,T) and
            T>R and
            payoff(D,C,T,S) and
            payoff(D,D,P,P) and
            R>P and
            P>S.
\end{lstlisting}

Since the autoformalization module does not impose constraints on the names of actions, this predicate also identifies which actions names correspond to specific actions.

\section{Methods}
\label{sec:methods}

\subsection{Parameters}

\begin{table}[htb]
\caption{Common experimental parameters.}
	\centering
    \footnotesize
	\begin{tabular}{lr}
		\toprule
             \textbf{Parameter} & \textbf{Value} \\
        \midrule
        \multirow{2}{*}{\texttt{LLM}} & \textit{\claude{}} \\
                                      & \textit{\gpt{}} \\ 
                                      \addlinespace
		\texttt{Temperature} & $1$ \\                                     
		\texttt{Maximum output tokens} & $2048$ \\
		\texttt{Maximum attempts number} & $5$ \\ 
        \bottomrule
      \end{tabular}
	
	\label{tab:exp-parameters-common}
\end{table}

\begin{table}[htb]
    \caption{Experiment-specific parameters. \textit{C} stands for a mode in which an agent plays against its clone with an opposite strategy, while \textit{RR} stands for round robin.}
    \centering
    \footnotesize
    \begin{tabular}{lcccc}
        \toprule
        \textbf{Parameter} & \textbf{Exp. 1} & \textbf{Exp. 2} & \textbf{Exp. 3} & \textbf{Exp. 4} \\ 
        \midrule
        \texttt{Games} & $55$ & $55$ & $5$ & $5$ \\ 
        \texttt{Agents in game} & $5$ & $5$ & $1$ & $1$ \\ 
        \texttt{Strategies} & $1$ & $1$ & $6$ & $6$ \\ 
        \texttt{Rounds} & $4$ & $4$ & $10$ & $10$ \\ 
        \texttt{Match maker} & C & C & RR & C \\ 
        \bottomrule
    \end{tabular}
    \label{tab:exp-parameters}
\end{table}

To evaluate the framework, we conducted four experiments: i) autoformalization of game scenarios with numerical (Exp. 1) and ii) non-numerical (Exp. 2) payoffs, iii) using the autoformalized game rules in a round-robin tournament (Exp.3), and iv) autoformalization of strategies (Exp. 4). Table~\ref{tab:exp-parameters-common} provides the experimental parameters shared across all experiments, while Table~\ref{tab:exp-parameters} details the parameters unique to each specific experiment.

\subsection{Dataset}

We used a dataset of 110 natural language game-theoretic scenarios, with and without numerical payoffs, based on an improved version of the dataset introduced in~\cite{mensfelt24a} (available in the repository). For both variants, the dataset included 5 scenarios per game employing common metaphors (e.g., two prisoners in the Prisoner’s Dilemma) for the five canonical games described in Section~\ref{sec:game-theory}. The remaining 50 scenarios—10 per game—used alternative descriptions that deliberately avoided these typical metaphors (e.g., two politicians competing in a campaign rather than two prisoners in the Prisoner’s Dilemma).

\subsection{Autoformalization of Interaction Descriptions (Numerical Payoffs)}

For each scenario, five agents autoformalized the game rules and subsequently played tournaments against clones of themselves, generating all four possible pairs of moves across four rounds. Each agent was allowed up to five attempts to produce syntactically correct code; agents that failed after five attempts were labeled as syntactically incorrect. Following the tournament, each agent’s individual and total payoffs were compared to the target values, with exact matches indicating semantic correctness.

\subsection{Autoformalization of Interaction Descriptions (Non-Numerical Payoffs)}

This experiment followed the same procedure as the numerical payoffs evaluation but used interaction scenarios without explicit numerical payoffs (e.g., ``If both remain silent, they receive minor sentences'' to describe mutual cooperation in the Prisoner's Dilemma). In these cases, the autoformalizing LLM was required to infer appropriate payoffs from the textual descriptions. As a ground truth payoff matrix could not be provided, semantic correctness was instead assessed through constraint checking (as described in Section~\ref{sec:solver}), verifying whether the generated payoff matrix satisfied the properties required for the corresponding game type (Section~\ref{sec:game-theory}).

\subsection{Axelrod's Tournament}

\begin{table*}[htb]
\caption{Strategies utilized in the evaluation.}
\centering
\small
\begin{tabularx}{\linewidth}{lX}
\toprule
\textbf{Strategy} & \textbf{Description} \\
\midrule
\textit{anti-default-move} & Always select the move that is the opposite of the default move. \\

\textit{anti-tit-for-tat} & Start with a default move. Then, select the move that is the opposite of the opponent's move in the previous round. \\

\textit{best-response} & Start with a default move. Then, select a move that would give you the highest payoff in response to the opponent's move in the previous round. \\

\textit{default-move} & Always select the default move. \\  

\textit{random} & Select one of the possible moves with uniform probability. \\  

\textit{tit-for-tat} & Start with a default move. Then, mirror the opponent's move in the previous round. \\ 
\bottomrule
\end{tabularx}
\label{tab:strategies}
\end{table*}

To demonstrate the framework’s applicability for analysing interaction scenarios, we selected five agents generated during the first experiment, one representing each of the five canonical games. For each game, we conducted a tournament inspired by Axelrod’s seminal tournament~\cite{axelrod1981evolution}. In this setup, six copies of each agent were created and assigned one of six strategies. Each agent competed against every other agent, including itself. The strategies are listed in Table~\ref{tab:strategies}.

The primary metric for evaluating strategy performance was the total payoff achieved by an agent using a particular strategy within each game. To enable cross-game comparisons of strategy effectiveness, total payoffs were normalized using the following formula:
{ 
\small \[
\mathrm{NormPayoff} = \frac{\mathrm{TotalPayoff} - \mathrm{MinTotalPayoff}}{\mathrm{MaxTotalPayoff} - \mathrm{MinTotalPayoff}}
\]
}

\noindent where {\small \texttt{TotalPayoff}} denotes the sum of payoffs accumulated by an agent across all rounds, and {\small \texttt{MinTotalPayoff}} and {\small \texttt{MaxTotalPayoff}} represent the minimum and maximum total payoffs achievable in that game, respectively.

\subsection{Autoformalization of Strategies}

For the autoformalization of strategies, we used the natural language description of the tit-for-tat strategy, along with its Prolog implementation, as an example for one-shot learning. The autoformalization module then received natural language descriptions of the remaining five strategies, similar to those shown in Table~\ref{tab:strategies}, to generate their representations.

Each autoformalized strategy was assigned to an agent equipped with predefined Prisoner's Dilemma rules. The agent then played four rounds against a clone using the anti-tit-for-tat strategy. As in the game autoformalization experiments, the agent's total payoff was compared against a target payoff specific to each strategy to assess semantic correctness. For the random strategy, correctness was additionally verified through manual inspection.

\section{Results and Discussion}
\label{sec:results}

\subsection{Autoformalization of Interaction Descriptions}

\begin{table}[htb]
\caption{Percentage of number of attempts at creating syntactically correct code for each model.}
\centering
\footnotesize
\begin{tabular}{l *{5}{S}}
\toprule
& \textbf{1 (\%)}   & \textbf{2 (\%)} & \textbf{3 (\%)} & \textbf{4 (\%)} & \textbf{5 (\%)} \\
\midrule
& \multicolumn{5}{c}{Numeric payoffs} \\
\midrule
\textbf{Claude 3.5 S} & 100.00 & 0.00 & 0.00 & 0.00 & 0.00 \\
\textbf{GPT-4o} & 96.77 & 2.87 & 0.00 & 0.00 & 0.36 \\
\midrule
& \multicolumn{5}{c}{Non-numeric payoffs} \\
\midrule
\textbf{Claude 3.5 S} & 100.00 & 0.00 & 0.00 & 0.00 & 0.00 \\
\textbf{GPT-4o} & 97.64 & 2.02 & 0.34 & 0.00 & 0.00 \\
\bottomrule
\end{tabular}
\label{tab:attempt_distribution}
\end{table}

Table~\ref{tab:attempt_distribution} presents the distribution of attempts required to generate syntactically correct code for both models. \claude{} achieved syntactically correct code on the first attempt in 100\% of cases, while \gpt{} did so in 97\% of cases.

The few errors observed in \gpt{} primarily involved non-Prolog comment characters or unescaped special characters in strings, most of which were corrected on the second attempt. These results demonstrate that providing the specific Prolog line that triggered the error in the feedback prompt significantly improves the LLM's ability to self-correct, compared to providing only an error message~\cite{mensfelt24a}. 

The only instance where \gpt{} failed to correct a syntactic error involved an ``\textit{Unexpected end of file}'' message. In this case, the feedback prompt did not include the problematic line itself, but only the generic error message, limiting the model's ability to identify and fix the issue.

\begin{table}[htb]
    \centering
    \caption{Syntactic, runtime, and semantic correctness for both models and types of scenario descriptions.}
    \footnotesize
    \resizebox{\columnwidth}{!}{
    \begin{tabular}{lccc}
        \toprule
        & \textbf{Syntactic (\%)} & \textbf{Runtime (\%)} & \textbf{Semantic (\%)} \\ 
        \midrule
        & \multicolumn{3}{c}{Numeric payoffs} \\ 
        \midrule
        \textbf{\claude{}} & 100.0 & 88.0 & 86.0 \\ 
        \textbf{\gpt{}} & 100.0 & 90.0 & 86.0 \\ 
        \midrule
        & \multicolumn{3}{c}{Non-numeric payoffs} \\ 
        \midrule
        \textbf{\claude{}} & 100.0 & 79.0 & 67.0 \\ 
        \textbf{\gpt{}} & 100.0 & 91.0 & 68.0 \\ 
        \bottomrule
    \end{tabular}
    }
    \label{tab:model_correctness}
\end{table}

The type of description had no influence on runtime errors in \gpt{} (runtime correctness of 90\% and 91\% for numeric and non-numeric payoffs, respectively, Table~\ref{tab:model_correctness}). However, it did have an influence in the case of \claude{}, with a 9 percentage point decrease to 79\% for non-numerical payoffs. A runtime error in this context means that the program was syntactically correct; however, while querying a predicate during runtime—most often to calculate payoffs—a predicate necessary to execute the query was missing. This highlights the need to extend the autoformalization feedback loop to the runtime stage and leverage the tracing mechanism of the solver to identify and create missing predicates.

\begin{table}[htb]
\centering
\caption{Correctness by game for \claude{} for game descriptions with numeric and non-numeric payoffs.}
\footnotesize
\begin{tabular}{l *{5}{S}}
\toprule
& \multicolumn{5}{c}{Numeric Payoffs} \\
\midrule
& \textbf{BS (\%)} & \textbf{HD (\%)} & \textbf{MP (\%)} & \textbf{PD (\%)} & \textbf{SH (\%)} \\
\midrule
\textbf{Syntactic} & 100.0 & 100.0 & 100.0 & 100.0 & 100.0 \\
\textbf{Runtime}   & 89.1  & 96.4  & 68.5  & 96.4  & 87.3  \\
\textbf{Semantic}  & 89.1  & 96.4  & 68.5  & 89.1  & 87.3  \\
\midrule
& \multicolumn{5}{c}{Non-Numeric Payoffs } \\
\midrule
& \textbf{BS (\%)} & \textbf{HD (\%)} & \textbf{MP (\%)} & \textbf{PD (\%)} & \textbf{SH (\%)} \\
\midrule
\textbf{Syntactic} & 100.0 & 100.0 & 100.0 & 100.0 & 100.0 \\
\textbf{Runtime}   & 82.1  & 85.7  & 74.0  & 70.7  & 83.1  \\
\textbf{Semantic}  & 82.1  & 77.8  & 34.2  & 64.0  & 83.1  \\
\bottomrule
\end{tabular}
\label{tab:correctness-claude}
\end{table}

\begin{table}[htb]
\centering
\caption{Correctness by game for \gpt{} for game descriptions with numeric and non-numeric payoffs.}
\footnotesize
\begin{tabular}{l *{5}{S}}
\toprule
& \multicolumn{5}{c}{Numeric Payoffs } \\
\midrule
& \textbf{BS (\%)} & \textbf{HD (\%)} & \textbf{MP (\%)} & \textbf{PD (\%)} & \textbf{SH (\%)} \\
\midrule
\textbf{Syntactic} & 98.3 & 100.0 & 100.0 & 100.0 & 100.0 \\
\textbf{Runtime}   & 83.1 & 89.1  & 96.4  & 89.1  & 90.9  \\
\textbf{Semantic}  & 79.7 & 89.1  & 90.9  & 85.5  & 83.6  \\
\midrule
& \multicolumn{5}{c}{Non-Numeric Payoffs (\%)} \\
\midrule
& \textbf{BS (\%)} & \textbf{HD (\%)} & \textbf{MP (\%)} & \textbf{PD (\%)} & \textbf{SH (\%)} \\
\midrule
\textbf{Syntactic} & 100.0 & 100.0 & 100.0 & 100.0 & 100.0 \\
\textbf{Runtime}   & 93.2  & 88.1  & 87.3  & 96.5  & 91.5  \\
\textbf{Semantic}  & 93.2  & 67.8  & 30.2  & 73.7  & 76.3  \\
\bottomrule
\end{tabular}
\label{tab:correctness-gpt4}
\end{table}

Semantic correctness was lower than runtime correctness for both models and description types (Table~\ref{tab:model_correctness}). This indicates that, although an agent was able to select moves and update its state during execution, the underlying semantics of the program were incorrect. For descriptions with numerical payoffs, semantic errors imply that the generated payoff matrix did not contain the required values. For descriptions without numerical payoffs, they indicate that the payoff matrix failed to satisfy the constraints associated with the intended game. Detailed syntactic, runtime, and semantic correctness results for different games are shown in Tables~\ref{tab:correctness-claude} and~\ref{tab:correctness-gpt4}.

Runtime and semantic correctness varied across models and games. Notably, Matching Pennies consistently exhibited low semantic correctness for non-numerical descriptions, with only 34.2\% correctness for \claude{} and 30.2\% for \gpt{}. This may be attributed to certain descriptions specifying only which player wins when choices match or mismatch, without explicitly stating that the other player loses—a crucial constraint of the game. Incorporating semantic evaluation into the autoformalization feedback loop may substantially improve correctness by addressing such issues.

\subsection{Axelrod's Tournament}

\begin{figure}[htb]
    \centering
    \includegraphics[width=\linewidth]{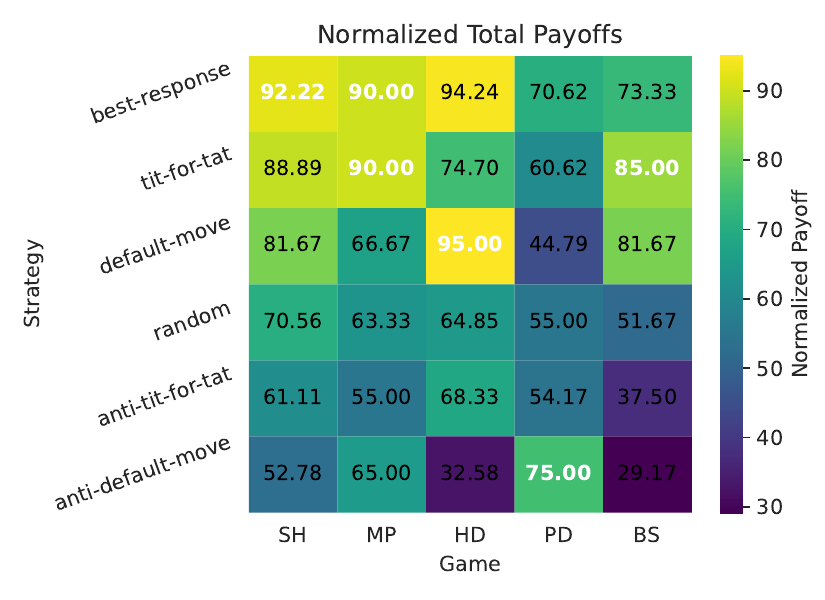}
\caption{Normalized total payoff for each strategy by game. Rows are ordered by the strategies with the highest average total payoff, and columns are arranged similarly based on the games. The total payoff for the winning strategy in each game is highlighted in white.}

    \label{fig:tournament}
\end{figure} 

Figure~\ref{fig:tournament} presents the normalized total payoffs for each strategy across all games. Overall, the best-response strategy proved the most effective on average, followed closely by tit-for-tat. Best-response achieved the highest scores in Stag Hunt and Matching Pennies (tied with tit-for-tat in the latter). However, in the Prisoner's Dilemma, it earned a lower payoff than the anti-default-move strategy (analogous to ``Defect''), as it opened with a ``Cooperate'' move.

These results demonstrate that agents equipped with autoformalized rules can successfully simulate strategic interactions and assess the relative performance of various strategies across different modelled scenarios.

\subsection{Autoformalization of Strategies}

\begin{table}[htb]
\caption{Percentage of semantic correctness for each autoformalized strategy.}
\centering
\footnotesize
\begin{tabular}{l SS}
\toprule
\textbf{Strategy} & \textbf{\claude{} (\%)} & \textbf{\gpt{} (\%)} \\
\midrule
anti-default-move & 100.00 & 100.00 \\
anti-tit-for-tat  & 100.00 & 100.00 \\
best-response     & 80.00  & 100.00 \\
default-move      & 100.00 & 100.00 \\
random            & 80.00  & 100.00 \\
\bottomrule
\end{tabular}
\label{tab:strategies-autoformalization}
\end{table}

Table~\ref{tab:strategies-autoformalization} presents the percentage of correctly autoformalized strategies. \gpt{} achieved 100\% correctness across all five strategies. \claude{} produced one incorrect autoformalization each for \textit{best-response} and \textit{random}, which are the most divergent from the \textit{tit-for-tat} strategy provided as an example in the prompt. Of these errors, one occurred during runtime and the other during semantic validation.

Overall, the high levels of correctness demonstrate that the GAMA framework can effectively autoformalize both game rules and gameplay strategies, including those with varying levels of similarity to the examples provided.

\section{Conclusions and Future Work}
\label{sec:conclusions}

We have proposed a framework for the autoformalization of interaction scenarios, focusing on those modelled as $2\times2$ simultaneous-move games. The framework supports autoformalization of both game rules and strategies, and enables simulation of tournaments based on the generated formal representations. Programs produced by the framework undergo three levels of validation: syntactic validation through solver consultation, runtime testing via gameplay, and semantic validation based on target payoffs or game-specific constraints. Experimental results demonstrated high syntactic and semantic accuracy across diverse interaction scenarios.

Future work will extend the framework beyond $2\times2$ simultaneous-move games. Our previous work~\cite{mensfelt24a} showed that generalization to structurally similar games, such as rock-paper-scissors and sequential versions of the Prisoner's Dilemma, could be addressed using the same solver. Generalization to a broader class of games, however, will require assembling a more diverse collection of in-context learning examples and developing retrieval mechanisms to select relevant examples for autoformalization in specific scenarios. Another promising direction involves expanding the feedback loop to incorporate runtime and semantic validation, further enhancing the correctness and robustness of the generated programs.

These extensions aim to establish a more general and reliable framework for generating reasoning modules, enabling decision-making agents to operate effectively across a wider range of games and interaction scenarios.


\begin{ack}
This work was supported by a Leverhulme Trust International Professorship Grant (LIP-2022-001). The first author would like to thank Bartosz Kosowski for the insightful discussions that contributed to the development of this research.
\end{ack}


\bibliography{references}

\begin{thebibliography}{39}
\providecommand{\natexlab}[1]{#1}
\providecommand{\url}[1]{\texttt{#1}}
\expandafter\ifx\csname urlstyle\endcsname\relax
  \providecommand{\doi}[1]{doi: #1}\else
  \providecommand{\doi}{doi: \begingroup \urlstyle{rm}\Url}\fi

\bibitem[Achiam et~al.(2023)Achiam, Adler, Agarwal, et~al.]{achiam2023gpt}
J.~Achiam, S.~Adler, S.~Agarwal, et~al.
\newblock {GPT}-4 technical report.
\newblock \emph{arXiv preprint arXiv:2303.08774}, 2023.

\bibitem[Akata et~al.(2023)Akata, Schulz, Coda-Forno, et~al.]{akata2023playing}
E.~Akata, L.~Schulz, J.~Coda-Forno, et~al.
\newblock Playing repeated games with large language models.
\newblock \emph{arXiv preprint arXiv:2305.16867}, 2023.

\bibitem[Anthropic(2024)]{claude35}
Anthropic.
\newblock Claude 3.5, 2024.
\newblock URL \url{https://www.anthropic.com/news/claude-3-5-sonnet}.

\bibitem[Axelrod and Hamilton(1981)]{axelrod1981evolution}
R.~Axelrod and W.~D. Hamilton.
\newblock The evolution of cooperation.
\newblock \emph{science}, 211\penalty0 (4489):\penalty0 1390--1396, 1981.

\bibitem[Chen et~al.(2023)Chen, Gandhi, Zhang, and Fan]{chen2023nl2tl}
Y.~Chen, R.~Gandhi, Y.~Zhang, and C.~Fan.
\newblock Nl2tl: Transforming natural languages to temporal logics using large
  language models.
\newblock In \emph{Proceedings of the 2023 Conference on Empirical Methods in
  Natural Language Processing}, pages 15880--15903, 2023.

\bibitem[Cosler et~al.(2023)Cosler, Hahn, Mendoza, Schmitt, and
  Trippel]{cosler2023nl2spec}
M.~Cosler, C.~Hahn, D.~Mendoza, F.~Schmitt, and C.~Trippel.
\newblock nl2spec: interactively translating unstructured natural language to
  temporal logics with large language models.
\newblock In \emph{International Conference on Computer Aided Verification},
  pages 383--396. Springer, 2023.

\bibitem[Fan et~al.(2023)Fan, Chen, Jin, and He]{fan2023can}
C.~Fan, J.~Chen, Y.~Jin, and H.~He.
\newblock Can large language models serve as rational players in game theory? a
  systematic analysis.
\newblock \emph{arXiv preprint arXiv:2312.05488}, 2023.

\bibitem[Feng et~al.(2023)Feng, Xu, Hao, Sharma, Shen, Zhao, and
  Chen]{feng2023language}
J.~Feng, R.~Xu, J.~Hao, H.~Sharma, Y.~Shen, D.~Zhao, and W.~Chen.
\newblock Language models can be logical solvers.
\newblock \emph{arXiv preprint arXiv:2311.06158}, 2023.

\bibitem[Genesereth et~al.(2005)Genesereth, Love, and Pell]{gdl}
M.~R. Genesereth, N.~Love, and B.~Pell.
\newblock General game playing: Overview of the {AAAI} competition.
\newblock \emph{{AI} Mag.}, 26\penalty0 (2):\penalty0 62--72, 2005.

\bibitem[Giacomo et~al.(2010)Giacomo, Lesp{\'{e}}rance, and Pearce]{sc-games1}
G.~D. Giacomo, Y.~Lesp{\'{e}}rance, and A.~R. Pearce.
\newblock Situation calculus based programs for representing and reasoning
  about game structures.
\newblock In F.~Lin, U.~Sattler, and M.~Truszczynski, editors, \emph{Principles
  of Knowledge Representation and Reasoning: Proceedings of the Twelfth
  International Conference, {KR} 2010, Toronto, Ontario, Canada, May 9-13,
  2010}. {AAAI} Press, 2010.

\bibitem[Gibbons(1992)]{Gibbons1992}
R.~Gibbons.
\newblock \emph{A Primer in Game Theory}.
\newblock Pearson Education Limited, Harlow, Essex, United Kingdom, 1992.

\bibitem[Gillioz et~al.(2020)Gillioz, Casas, Mugellini, and
  Khaled]{gillioz2020transformers}
A.~Gillioz, J.~Casas, E.~Mugellini, and O.~A. Khaled.
\newblock Overview of the transformer-based models for nlp tasks.
\newblock In \emph{2020 15th Conference on Computer Science and Information
  Systems (FedCSIS)}, pages 179--183, 2020.
\newblock \doi{10.15439/2020F20}.

\bibitem[Gou et~al.(2024)Gou, Shao, Gong, et~al.]{gou2024critic}
Z.~Gou, Z.~Shao, Y.~Gong, et~al.
\newblock Critic: Large language models can self-correct with tool-interactive
  critiquing.
\newblock \emph{arXiv preprint arXiv:2305.11738}, 2024.

\bibitem[Guo(2023)]{guo2023gpt}
F.~Guo.
\newblock {GPT} agents in game theory experiments.
\newblock \emph{arXiv preprint arXiv:2305.05516}, 2023.

\bibitem[He-Yueya et~al.(2023)He-Yueya, Poesia, Wang, and
  Goodman]{he2023solving}
J.~He-Yueya, G.~Poesia, R.~E. Wang, and N.~D. Goodman.
\newblock Solving math word problems by combining language models with symbolic
  solvers.
\newblock \emph{arXiv preprint arXiv:2304.09102}, 2023.

\bibitem[Imani et~al.(2023)Imani, Du, and Shrivastava]{imani2023mathprompter}
S.~Imani, L.~Du, and H.~Shrivastava.
\newblock Mathprompter: Mathematical reasoning using large language models.
\newblock \emph{arXiv preprint arXiv:2303.05398}, 2023.

\bibitem[Jiang et~al.(2022)Jiang, Welleck, Zhou, Li, Liu, Jamnik, Lacroix, Wu,
  and Lample]{jiang2022draft}
A.~Q. Jiang, S.~Welleck, J.~P. Zhou, W.~Li, J.~Liu, M.~Jamnik, T.~Lacroix,
  Y.~Wu, and G.~Lample.
\newblock Draft, sketch, and prove: Guiding formal theorem provers with
  informal proofs.
\newblock \emph{arXiv preprint arXiv:2210.12283}, 2022.

\bibitem[Lesperance et~al.(2024)Lesperance, De~Giacomo, Rostamigiv, and
  Khan]{sc-games2}
Y.~Lesperance, G.~De~Giacomo, M.~Rostamigiv, and S.~M. Khan.
\newblock Abstraction of situation calculus concurrent game structures.
\newblock \emph{Proceedings of the AAAI Conference on Artificial Intelligence},
  38\penalty0 (9):\penalty0 10624--10634, Mar. 2024.
\newblock \doi{10.1609/aaai.v38i9.28933}.
\newblock URL \url{https://ojs.aaai.org/index.php/AAAI/article/view/28933}.

\bibitem[Lor{\`e} and Heydari(2023)]{lore2023strategic}
N.~Lor{\`e} and B.~Heydari.
\newblock Strategic behavior of large language models: Game structure vs.
  contextual framing.
\newblock \emph{arXiv preprint arXiv:2309.05898}, 2023.

\bibitem[Love et~al.(2006)Love, Hinrichs, Haley, Schkufza, and
  Genesereth]{gdl-orig}
N.~Love, T.~Hinrichs, D.~Haley, E.~Schkufza, and M.~Genesereth.
\newblock {General Game Playing: Game Description Language Specification}.
\newblock Technical Report LG–2006–01, Stanford University, 2006.

\bibitem[Madaan et~al.(2023)Madaan, Tandon, Gupta,
  et~al.]{madaan2023selfrefine}
A.~Madaan, N.~Tandon, P.~Gupta, et~al.
\newblock Self-refine: Iterative refinement with self-feedback.
\newblock \emph{arXiv preprint arXiv:2303.17651}, 2023.

\bibitem[McCarthy and Hayes(1981)]{sc}
J.~McCarthy and P.~Hayes.
\newblock Some philosophical problems from the standpoint of artificial
  intelligence.
\newblock In B.~L. Webber and N.~J. Nilsson, editors, \emph{Readings in
  Artificial Intelligence}, pages 431--450. Morgan Kaufmann, 1981.

\bibitem[Mensfelt et~al.(2024)Mensfelt, Stathis, and Tencsenyi]{mensfelt24a}
A.~Mensfelt, K.~Stathis, and V.~Tencsenyi.
\newblock Autoformalization of {G}ame {D}escriptions using {L}arge {L}anguage
  {M}odels.
\newblock In \emph{1st {I}nternational {W}orkshop on {N}ext-{G}eneration
  {L}anguage {M}odels for {K}nowledge {R}epresentation and {R}easoning}, Hanoi,
  Vietnam, 2024.
\newblock URL \url{https://arxiv.org/abs/2409.12300}.

\bibitem[{OpenAI}(2024)]{gpt4o2024}
{OpenAI}.
\newblock {GPT}-4o.
\newblock \url{https://openai.com/index/hello-gpt-4o/}, 2024.
\newblock URL \url{https://openai.com/index/hello-gpt-4o/}.
\newblock Accessed: 25/07/2024.

\bibitem[Osborne and Rubinstein(1994)]{OsborneRubinstein1994}
M.~Osborne and A.~Rubinstein.
\newblock \emph{A Course in Game Theory}.
\newblock The MIT Press, 1994.

\bibitem[Pan et~al.(2023)Pan, Albalak, Wang, and Wang]{pan2023logic}
L.~Pan, A.~Albalak, X.~Wang, and W.~Wang.
\newblock Logic-lm: Empowering large language models with symbolic solvers for
  faithful logical reasoning.
\newblock In \emph{Findings of the Association for Computational Linguistics:
  EMNLP 2023}, pages 3806--3824, 2023.

\bibitem[Qiu et~al.(2020)Qiu, Sun, Xu, Shao, Dai, and Huang]{Qiu2020}
X.~Qiu, T.~Sun, Y.~Xu, Y.~Shao, N.~Dai, and X.~Huang.
\newblock Pre-trained models for natural language processing: A survey.
\newblock \emph{Science China Technological Sciences}, 63:\penalty0 1872--1897,
  2020.
\newblock \doi{10.1007/s11431-020-1647-3}.

\bibitem[Rasmusen(2006)]{Rasmusen2004introtogametheory}
E.~Rasmusen.
\newblock \emph{Games and information an introduction to game theory}.
\newblock Blackwell, 2006.

\bibitem[Renze and Guven(2024)]{renze2024self}
M.~Renze and E.~Guven.
\newblock Self-reflection in llm agents: Effects on problem-solving
  performance.
\newblock \emph{arXiv preprint arXiv:2405.06682}, 2024.

\bibitem[Scherl and Levesque(2003)]{sc-kbf}
R.~B. Scherl and H.~J. Levesque.
\newblock Knowledge, action, and the frame problem.
\newblock \emph{Artif. Intell.}, 144\penalty0 (1-2):\penalty0 1--39, 2003.
\newblock \doi{10.1016/S0004-3702(02)00365-X}.
\newblock URL \url{https://doi.org/10.1016/S0004-3702(02)00365-X}.

\bibitem[Schiffel and Thielscher(2011)]{Schiffel_Thielscher_2011}
S.~Schiffel and M.~Thielscher.
\newblock Reasoning about general games described in gdl-ii.
\newblock \emph{Proceedings of the AAAI Conference on Artificial Intelligence},
  25\penalty0 (1):\penalty0 846--851, Aug. 2011.
\newblock \doi{10.1609/aaai.v25i1.7944}.

\bibitem[Shinn et~al.(2023)Shinn, Cassano, Berman, et~al.]{shinn2023reflexion}
N.~Shinn, F.~Cassano, E.~Berman, et~al.
\newblock Reflexion: Language agents with verbal reinforcement learning.
\newblock \emph{arXiv preprint arXiv:2303.11366}, 2023.

\bibitem[Thielscher(2010)]{gdl-ii}
M.~Thielscher.
\newblock A general game description language for incomplete information games.
\newblock In \emph{Proceedings of the Twenty-Fourth AAAI Conference on
  Artificial Intelligence}, AAAI'10, page 994–999. AAAI Press, 2010.

\bibitem[Thielscher(2011)]{gdl-univ}
M.~Thielscher.
\newblock The general game playing description language is universal.
\newblock In T.~Walsh, editor, \emph{{IJCAI} 2011, Proceedings of the 22nd
  International Joint Conference on Artificial Intelligence, Barcelona,
  Catalonia, Spain, July 16-22, 2011}, pages 1107--1112. {IJCAI/AAAI}, 2011.

\bibitem[Wang and
  Zhao(2024)]{wang2024rupbenchbenchmarkingreasoningperturbations}
Y.~Wang and Y.~Zhao.
\newblock Rupbench: Benchmarking reasoning under perturbations for robustness
  evaluation in large language models.
\newblock \emph{arXiv preprint arXiv:2406.11020}, 2024.

\bibitem[Wielemaker et~al.(2012)Wielemaker, Schrijvers, Triska, and
  Lager]{wielemaker2012swi}
J.~Wielemaker, T.~Schrijvers, M.~Triska, and T.~Lager.
\newblock Swi-prolog.
\newblock \emph{Theory and Practice of Logic Programming}, 12\penalty0
  (1-2):\penalty0 67--96, 2012.

\bibitem[Wu et~al.(2022)Wu, Jiang, Li, Rabe, Staats, Jamnik, and
  Szegedy]{wu2022autoformalization}
Y.~Wu, A.~Q. Jiang, W.~Li, M.~Rabe, C.~Staats, M.~Jamnik, and C.~Szegedy.
\newblock Autoformalization with large language models.
\newblock \emph{Advances in Neural Information Processing Systems},
  35:\penalty0 32353--32368, 2022.

\bibitem[Yang et~al.(2023)Yang, Ishay, and Lee]{yang2023coupling}
Z.~Yang, A.~Ishay, and J.~Lee.
\newblock Coupling large language models with logic programming for robust and
  general reasoning from text.
\newblock \emph{arXiv preprint arXiv:2307.07696}, 2023.

\bibitem[Zhao et~al.(2023)Zhao, Zhou, Li, et~al.]{zhao2023survey}
W.~X. Zhao, K.~Zhou, J.~Li, et~al.
\newblock A survey of large language models.
\newblock \emph{arXiv preprint arXiv:2303.18223}, 2023.

\end{thebibliography}


\end{document}